%% file: root.tex
\documentclass[letterpaper, 10 pt, conference]{ieeeconf}  

\IEEEoverridecommandlockouts                              

\title{\LARGE \bf
SIGMA: Sheaf-Informed Geometric Multi-Agent Pathfinding
}

\author{Shuhao Liao$^{1}$, Weihang Xia$^{2}$, Yuhong Cao$^{3}$, Weiheng Dai$^{3}$,\\ Chengyang He$^{3}$, Wenjun Wu$^{1,*}$, Guillaume Sartoretti$^{3}$%
\thanks{$^{1}$Hangzhou International Innovation Institute, Beihang University, China}%
\thanks{$^{2}$CoreControl Inc, Hangzhou, China}
\thanks{$^{3}$Department of Mechanical Engineering, National University of Singapore, Singapore}
\thanks{*Corresponding author: Wenjun Wu (wwj09315@buaa.edu.cn).}
}

\usepackage{amssymb}
\usepackage{amsmath}
\usepackage{graphicx}
\usepackage{multirow}
\usepackage{booktabs}
\usepackage{tabularx}
\usepackage{subcaption}
\usepackage{cite}
\usepackage{hyperref}
\begin{document}

\maketitle
\thispagestyle{empty}
\pagestyle{empty}

\begin{abstract}

The Multi-Agent Path Finding (MAPF) problem aims to determine the shortest and collision-free paths for multiple agents in a known, potentially obstacle-ridden environment. It is the core challenge for robotic deployments in large-scale logistics and transportation. Decentralized learning-based approaches have shown great potential for addressing the MAPF problems, offering more reactive and scalable solutions. However, existing learning-based MAPF methods usually rely on agents making decisions based on a limited field of view (FOV), resulting in short-sighted policies and inefficient cooperation in complex scenarios. There, a critical challenge is to achieve consensus on potential movements between agents based on limited observations and communications. To tackle this challenge, we introduce a new framework that applies sheaf theory to decentralized deep reinforcement learning, enabling agents to learn geometric cross-dependencies between each other through local consensus and utilize them for tightly cooperative decision-making. In particular, sheaf theory provides a mathematical proof of conditions for achieving global consensus through local observation. Inspired by this, we incorporate a neural network to approximately model the consensus in latent space based on sheaf theory and train it through self-supervised learning. During the task, in addition to normal features for MAPF as in previous works, each agent distributedly reasons about a learned consensus feature, leading to efficient cooperation on pathfinding and collision avoidance. As a result, our proposed method demonstrates significant improvements over state-of-the-art learning-based MAPF planners, especially in relatively large and complex scenarios, demonstrating its superiority over baselines in various simulations and real-world robot experiments. The code is available at \url{https://github.com/marmotlab/SIGMA}

\end{abstract}
\section{INTRODUCTION}

As intelligent robots advance, the application of large-scale Multi-Agent Path Finding (MAPF) has become increasingly important in scenarios such as warehouse automation, airport management, and robotic fleets~\cite{honig2019persistent, rekik2019multi, li2021lifelong, he2024social}. MAPF involves planning collision-free paths for multiple agents from their start positions to designated goals. This NP-hard problem presents significant challenges in scalability and computational efficiency due to the exponential growth of complexity with respect to the number of agents. 

Recently, the MAPF community has started looking to Multi-Agent Reinforcement Learning (MARL) to generate fast and scalable solutions~\cite{primal2,scrimp,g2rl}. Moreover, MARL has gained significant traction in multi-robot systems, where agents collaborate in decentralized settings to achieve global objectives~\cite{yu2023esp,yu2024adaptaug,yu2024leveraging,yu2021swarm,feng2024safe,feng2023mact}. These learning-based approaches rely on decentralized planning under partial observability (i.e., each agent only observes its nearby environment, usually 11$\times$11 grid world), reducing computational complexity and enabling the network to tackle large-scale scenarios effectively~\cite{feng2024hierarchical}. However, the solutions generated by these learning-based methods are usually suboptimal, since they restrict the information available to agents, hindering their ability to avoid local minima and perform delicate joint behaviors. To improve the performance of learning-based MAPF planners, recent methods tried to augment the information available to agents by incorporating expert paths, designing communication schemes, or providing global map encodings~\cite{primal, dhc, alpha}. These methods show significant improvements over previous learning-based methods, but there is still a remarkable performance gap between learning-based solutions and centralized optimization-based solutions~\cite{odrm}.

\begin{figure}[t]
  \centering
  \includegraphics[scale=0.22]{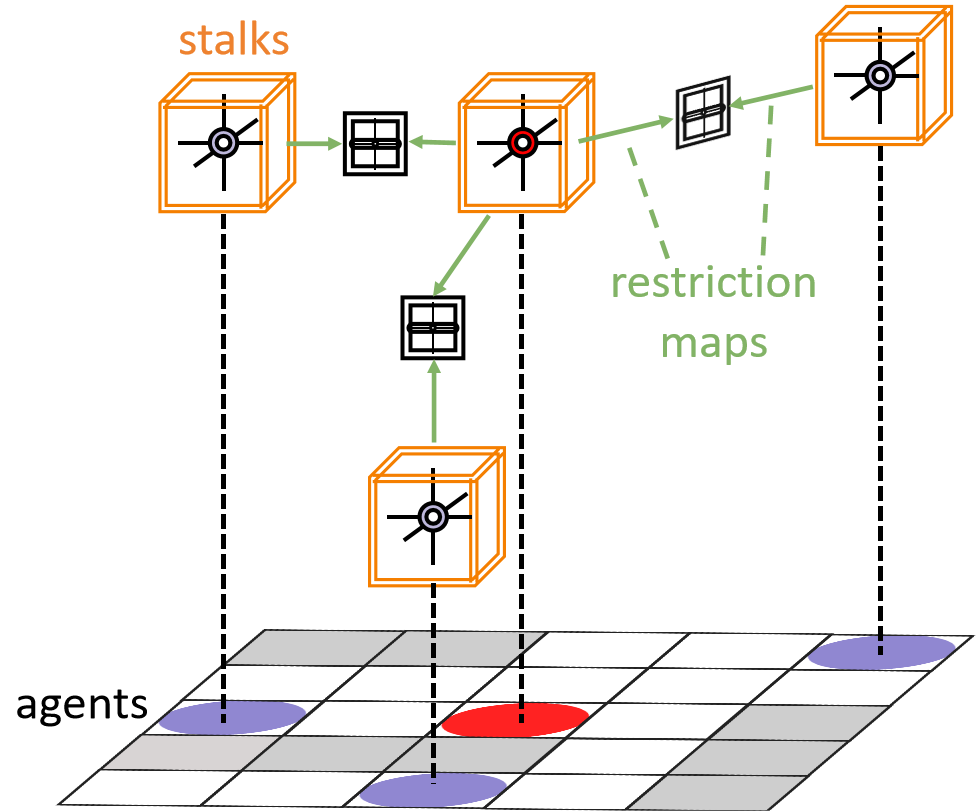}
  \caption{A sheaf structure in MAPF, where the sheaf structure provides multilayer views of system structure, stalks collate high-dimensional data associated with agents, and restriction maps describe complex relationships between different agents. This sheaf structure aids agents in implementing consensus.}
  \label{pic1}
  \vspace{-5mm}
\end{figure}

This work builds upon the observation that existing learning-based planners typically lack the ability to let agents reason about and reach an explicit \textit{consensus} (i.e., a general agreement between agents on their future movements). Such consensus is naturally achieved in centralized methods, which significantly helps agents avoid blockage and deadlock, leading to a better success rate and shorter makespan. However, achieving consensus is non-trivial in decentralized methods, as each agent makes its decision individually.
As a result, current state-of-the-art learning-based methods still struggle with dense environments that require tight cooperation among agents.

To address this problem, we introduce SIGMA, a novel sheaf-informed MAPF planner that explicitly trains consensus by learning the underlying geometric cross-dependencies between agents. 
To the best of our knowledge, this work is the first learning-based method in MAPF that helps agents explicitly achieve consensus among themselves.
Our method enhances the capability of the trained planner for modeling complex potential interactions between agents. In particular, we integrate \textit{sheaf theory}~\cite{hansen2019toward} into decentralized deep reinforcement learning, enabling agents to learn consensus/global consistency by modeling geometric cross-dependencies between each other, where these geometric cross-dependencies represent the consensus. Sheaf theory broadens the concept of graphs and studies the global consistency of high dimensional data. Particularly, it provides mathematical proof of conditions for achieving global consensus through local observation. Inspired by sheaf theory, we incorporate a neural network to approximately model the consensus in latent space based on sheaf theory and train it through self-supervised learning. By doing so, in addition to normal features for MAPF as in previous works, each agent distributedly reasons about a learned consensus feature for consensus-aware path-finding and collision avoidance.  We present exhaustive numerical comparisons with existing conventional and learning-based planners, which show that SIGMA outperforms state-of-the-art learning-based MAPF planners. Notably, SIGMA excels in large-scale scenarios with larger team sizes, where it significantly surpasses existing methods.



\section{RELATED WORK}

\subsection{Deep Reinforcement Learning-based MAPF}

Recent years have seen a growing interest in solving MAPF problems using MARL. The pioneering work, PRIMAL, introduced a combination of RL and imitation learning to plan paths through fully decentralized policies within a partially observable environment~\cite{primal}. PRIMAL utilized the Asynchronous Advantage Actor-Critic algorithm as underlying RL algoritihm, with all agents sharing the same parameters, while imitation learning was based on behavior cloning from data generated by the ODrM* planner. This approach was later extended in PRIMAL2 to address lifelong MAPF scenarios, incorporating learned conventions to enhance cooperation among agents, particularly in highly structured environments~\cite{primal2}. Subsequent research has explored communication learning as a promising approach to further enhance solution quality. For instance, works like MAGAT and DHC~\cite{magat, dhc} introduced Graph Neural Networks~\cite{peng2024graphrare} for communication learning, where each agent is treated as a node, and decisions are made based on aggregated information from neighboring agents. DCC, on the other hand, developed a selective communication strategy that determines whether an agent's decision should be influenced by its neighbors~\cite{dcc}. Moreover, PICO integrated planning priorities from a classical coupled planner into an ad-hoc communication topology, aiming to produce policies that reduce collisions~\cite{pico}. More recently, SCRIMP introduced a scalable method where agents learn from small FOV with a modified transformer for communication, improving performance in dense scenarios~\cite{scrimp}. Similarly, ALPHA combined local and global information, using a Graph Transformer to enhance decision-making and cooperation, addressing the limitations of limited FOV~\cite{alpha}. However, these methods often face challenges of scalability and complexity, struggling to handle instances with agent-dense environments.

\subsection{Sheaf Applications}
Sheaf theory addresses the local-to-global problem in multi-agent systems by allowing the coherent integration of local data into global structures~\cite{sheafori}. Grothendieck then extended its application in algebraic geometry through scheme theory, enabling the handling of complex structures like singularities~\cite{grothendieck1955general}. In distributed systems, sheaf theory provides a framework for consensus on complex data structures~\cite{hansen2020laplacians}. In signal processing, it manages distributed data with complex dependencies, and in network communication, it optimizes information flow and reduces communication costs~\cite{robinson2014topological}. In network science, sheaves provide enhanced descriptions of network structures, capturing the nature of relationships between nodes. Within this framework, opinion dynamics uses discourse sheaves to model how opinions evolve and interact within social networks~\cite{hansen2021opinion}. Additionally, Bodnar proposed neural sheaf diffusion to learn sheaf laplacians from lower-order data, providing a novel topological perspective on heterophily and oversmoothing in GNNs~\cite{bodnar2022neural,10872817,peng2023clgt}.

\begin{figure*}[t]
  \centering
  \includegraphics[scale=0.74]{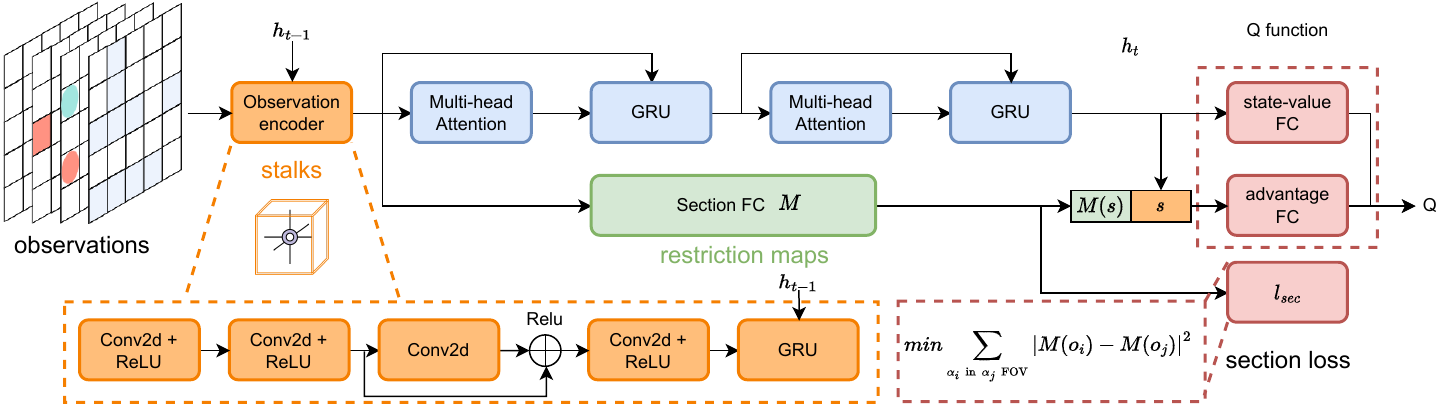}
  \caption{Network structure of SIGMA. The observation encoder encodes observations in stalks (orange), the section FC learns restriction maps (green), and \(M(s)\) is included in the advantage function to enhance action evaluation. A global section loss (red) is then integrated into the network updating to align the policy with the sheaf structure.}
  \label{framework}
\end{figure*}

\section{MAPF AS AN RL PROBLEM}

\subsection{MAPF Problem Statement}


The \textit{classical} MAPF problem considers a set of agents $N=\left\{\alpha_1, \ldots, \alpha_n\right\}$ and an undirected graph \(\mathbf{G} = (\mathbf{V}, \mathbf{E})\), where \(\mathbf{V}\) represents the set of vertices and \(\mathbf{E}\) represents the set of edges. Each agent \(i\) is assigned a distinct start vertex (\(\xi_i \in \mathbf{V}\)) and a distinct goal vertex (\(g_i \in \mathbf{V}\)). Time is discretized into uniform steps. At each time step \(t = 0, 1, 2, \ldots\), an agent has the option to either move to an adjacent vertex or remain stationary at its current vertex. A path for agent \(\alpha_i\) is defined as a sequence of vertices, either adjacent (indicating movement) or identical (indicating waiting), starting from the agent's start vertex \(\xi_i\) and ending at its goal vertex \(g_i\). Collisions between agents are categorized as either vertex collisions, which occur when two agents \(\alpha_i\) and \(\alpha_j\) occupy the same vertex \(v\) at the same time \(t\), or edge collisions, which occur when two agents \(\alpha_i\) and \(\alpha_j\) simultaneously traverse the same edge \((u, v)\) in opposite directions at time \(t\). A valid solution to the MAPF problem is a set of collision-free paths, one for each agent. The optimality of the solution is typically evaluated by the sum of the arrival times of all agents at their respective goal vertices.

\subsection{RL Environment Setup}
Remaining consistent with the standard MAPF problem, we use the following setup: the map is a 2D discrete 4-connectivity grid world,  where each grid is a vertex connected to its neighbors by edges, and agents can move to the free cell adjacent to their location or stay idle at each time step. An episode terminates when all agents are on their goals at the end of a time step (success) or when the number of time steps reaches the pre-defined limit (failure). We utilize the room-like map generator proposed by ALPHA. The generated room maps contain corridors of varying widths, which closely resemble real offices, warehouses, and other environments. Unlike the temporal impact of loose obstacles in random maps, continuous obstacles in such highly structured maps may significantly affect current decision-making even across substantial distances.

\subsubsection{Observation}
In our settings, each agent can only partially observe the environment limited by the size of the FOV \(\ell \times \ell\), where \(\ell\) is smaller than the total environment size \(m\). To ensure the agent remains centered within its FOV, \(\ell\) is selected as an odd number. The observation data is organized into two primary channels: the first channel is a binary matrix that represents obstacles within the FOV, and the second channel is another binary matrix that indicates the positions of other agents when they fall within the FOV. Additionally, the input to our model includes four heuristic channels, each corresponding to one of the possible movement directions: Up, Down, Left, and Right. These heuristic channels share the same dimensions as the FOV, and each cell is marked as 1 if taking the associated action would move the agent closer to its goal from that location. By incorporating these heuristic channels, the agent can infer the best direction to move without the need to explicitly include the goal location in the input~\cite{dhc}.

\subsubsection{Action Space}
In our grid-world environment, agents operate within a discrete action space. At each time step, an agent can choose to move to one of the adjacent grid cells or remain stationary. We do not consider diagonal movements thus each agent has a total of five possible actions. During training or execution, agents may occasionally select invalid actions, such as moving into an obstacle or causing a collision with another agent. To handle such situations, invalid actions are not filtered out; instead, if an invalid action occurs, the agent and any involved agents are recursively returned to their previous states until no collisions remain~\cite{dhc}.

\subsubsection{Reward}

We follow the DHC setup~\cite{dhc}, assigning the same penalty for each agent's movement and for staying away from the goal. The same reward setting ensures a fairer comparison. Our reward structure is shown in Table~\ref{tab:rewards}.


\begin{table}[ht]
    \centering
    \caption{Reward Structure}
    \label{tab:rewards}
    \begin{tabular}{c||c}
        \hline
        \textbf{Actions} & \textbf{Reward} \\
        \hline
        Move (Up/Down/Left/Right) & $-0.075$ \\
        \hline
        Stay (on goal, away goal) & $0,-0.075$ \\
        \hline
        Collision (obstacle/agents) & $-0.5$ \\
        \hline
        Finish & $3$ \\
        \hline
    \end{tabular}
\end{table}

\section{LEARNING TECHNIQUES}
\subsection{Dynamic Agent Graph}

\begin{figure}[thpb]
  \centering
  \includegraphics[scale=0.07]{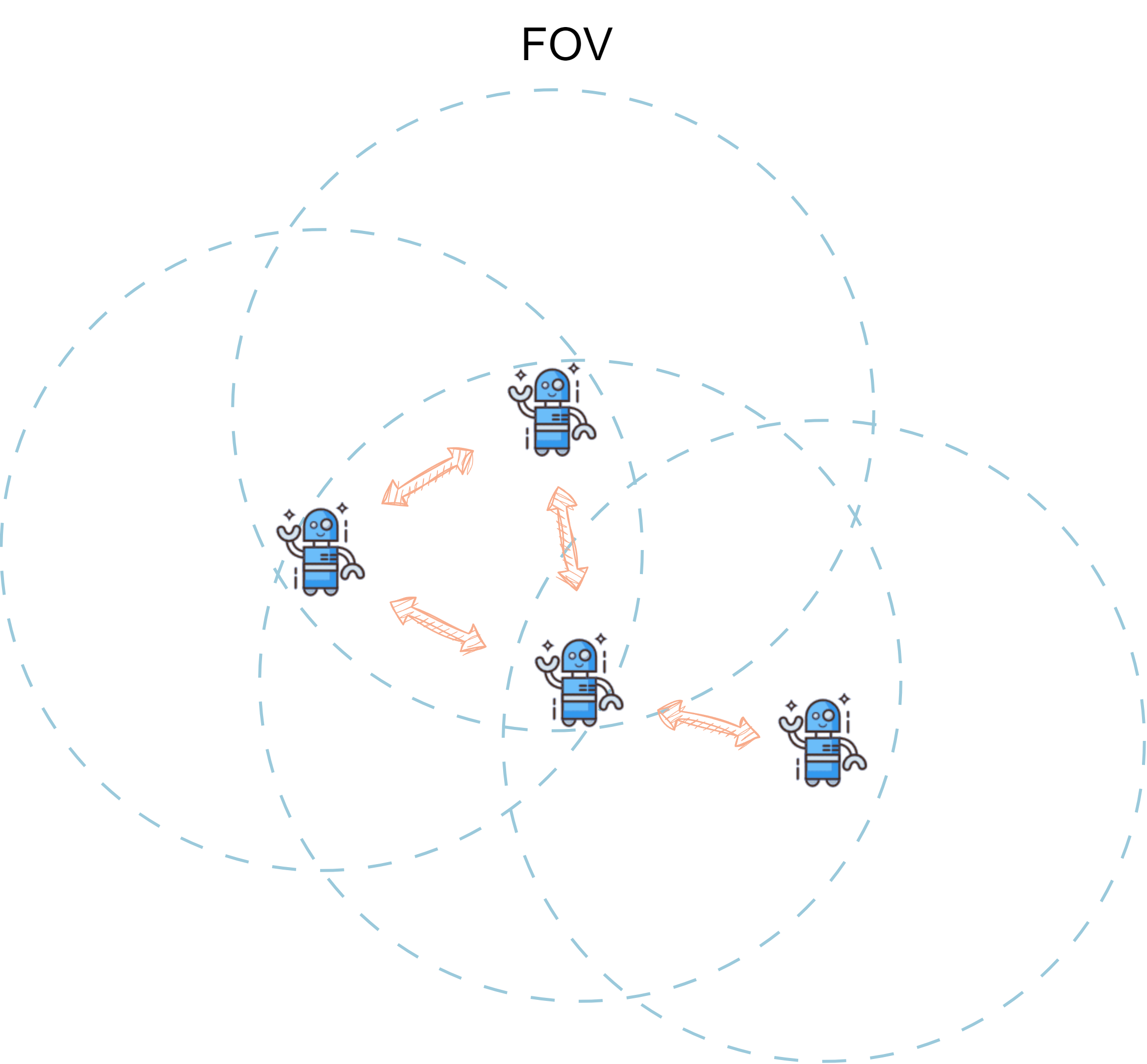}
  \caption{Dynamic agent graph shows dynamic connections among homogeneous agents, with edges representing mutual visibility within each other's FOV.}
  \label{agent graph}
\end{figure}

We define a dynamic agent graph $G=(V, E)$ to represent the relationships between agents based on their FOV, as illustrated in Figure~\ref{agent graph}. Nodes $V$ represent $n$ agent $\alpha_i, i=(1,2,...,n)$. An edge $e \in E$ is established between two agents if they are within each other's FOV, indicating that they can potentially interact or need to consider each other's presence while planning their paths. This dynamic graph reflects the changing visibility and proximity of agents as they move through the environment, making it crucial for coordinating their actions and avoiding collisions. 

\subsection{Cellular Sheaf in MAPF}
A cellular sheaf is an algebraic-topological structure associated with a graph that attaches spaces of data to nodes and edges~\cite{snn}. To be precise, a cellular sheaf $(G, \mathcal{F})$ on a dynamic agent graph $G=(V, E)$ consists of:
\begin{itemize}
\item  A vector space $\mathcal{F}(v)$ for each $v \in V$,
\item  A vector space $\mathcal{F}(e)$ for each $e \in E$,
\item  A linear map $\mathcal{F}_{v \unlhd e}: \mathcal{F}(v) \rightarrow \mathcal{F}(e)$ for each incident node-edge pair $v \unlhd e$.
\end{itemize}

According to the sheaf theory, vector spaces is termed as \textit{stalks} and the linear map as \textit{restriction map}~\cite{Sheaftheory}. The stalk originates from the analogy where a sheaf in the agricultural sense is a collection of stalks of grain bound together by twine; similarly, in mathematical terminology, a cellular sheaf on a graph is a collection of stalks of data bound together by restriction maps.


Here we focus on the concept of consensus and help readers to understand how it works in MAPF. Agents achieve global consensus via local observations, where the local observations of agents are in node stalks, and the model dependences between agents are in edge stalks. In our case,
agents reach consensus when their independent observations map to consistent features via learned restriction maps.
Specifically, if for any two agents $ v $ and $ u $ with observation vectors $ x_v $ and $ x_u $ agree on the edge $ e $, the condition $ \mathcal{F}_{v \unlhd e} x_v = \mathcal{F}_{u \unlhd e} x_u $ should be  satisfied. Here, the subspace of direct sum of all the node stalks that satisfy this condition is referred to as the space of global sections $\Gamma(G, \mathcal{F})$ \cite{globalsection}. We regard elements in the space of global sections as representatives of consensus, where the entire agents exhibit no-contradiction behavior in how to map the observations of agents across the graph to global sections. In our work, our target is to utilize self-supervising learning to train a neural networ to model the space of global sections, achieving consensus to avoid congestion and crowding among agents.

\subsection{Sheaf-Informed DQN}

As shown in Figure~\ref{framework}, we use a observation encoder to encode the observations into stalks (orange). Since MAPF agents are homogeneous, meaning they have identical characteristics, their corresponding restriction maps are also identical. Thus, we denote the same restriction maps (green) for all agents as $ M $, which is learned through the section FC's training process. Additionally, we incorporate the global section loss(red) directly into the network updating, ensuring that the learned policy respects the underlying sheaf structure and maintains consistency across the observation space.

DQN learns the action value function using neural networks. To incorporate the advantage and value functions, we can define the advantage function \( A(s, a) \) and the value function \( V(s) \), where $Q(s, a) = V(s) + A(s, a)$.
The agents access the current states \( s_t \in \mathcal{S} \) and selects an action \( a_t \in \mathcal{A} \) according to a policy \( \pi \) at each time step \( t \). The agent's objective is to maximize the expectation of the discounted total return \( R_t = r_t + \gamma r_{t+1} + \gamma^2 r_{t+2} + \ldots \), where \( r_t \) is the reward received at time \( t \).

Q-Learning utilizes an action value function for policy \( \pi \) as \( Q^\pi(s, a) = \mathbb{E}[R_t \mid s_t = s, a_t = a] \) and can be recursively defined by $Q^\pi(s, a) = \mathbb{E}_{s'}[r + \gamma \mathbb{E}_{a' \sim \pi}[Q^\pi(s', a')]]$
The optimal action value, \( Q^*(s, a) = \max_\pi Q^\pi(s, a) \), satisfies the Bellman optimality equation $Q^*(s, a) = \mathbb{E}_{s'}[r + \gamma \max_{a'} Q^*(s', a') \mid s, a]$. The optimal policy is trained by minimizing the loss $\mathcal{L_Q}$. Here, the parameters of the target network are updated periodically. In partially observable environments, agents generally need to condition on an state-action history $\mathcal{L_Q} = \mathbb{E}_{(s, a, r, s')}[(Q(s, a) - y)^2]$, 
where $y = r + \gamma \max_{a'} Q(s', a')$.

\begin{figure}[thpb]
  \centering
  \includegraphics[scale=0.1]{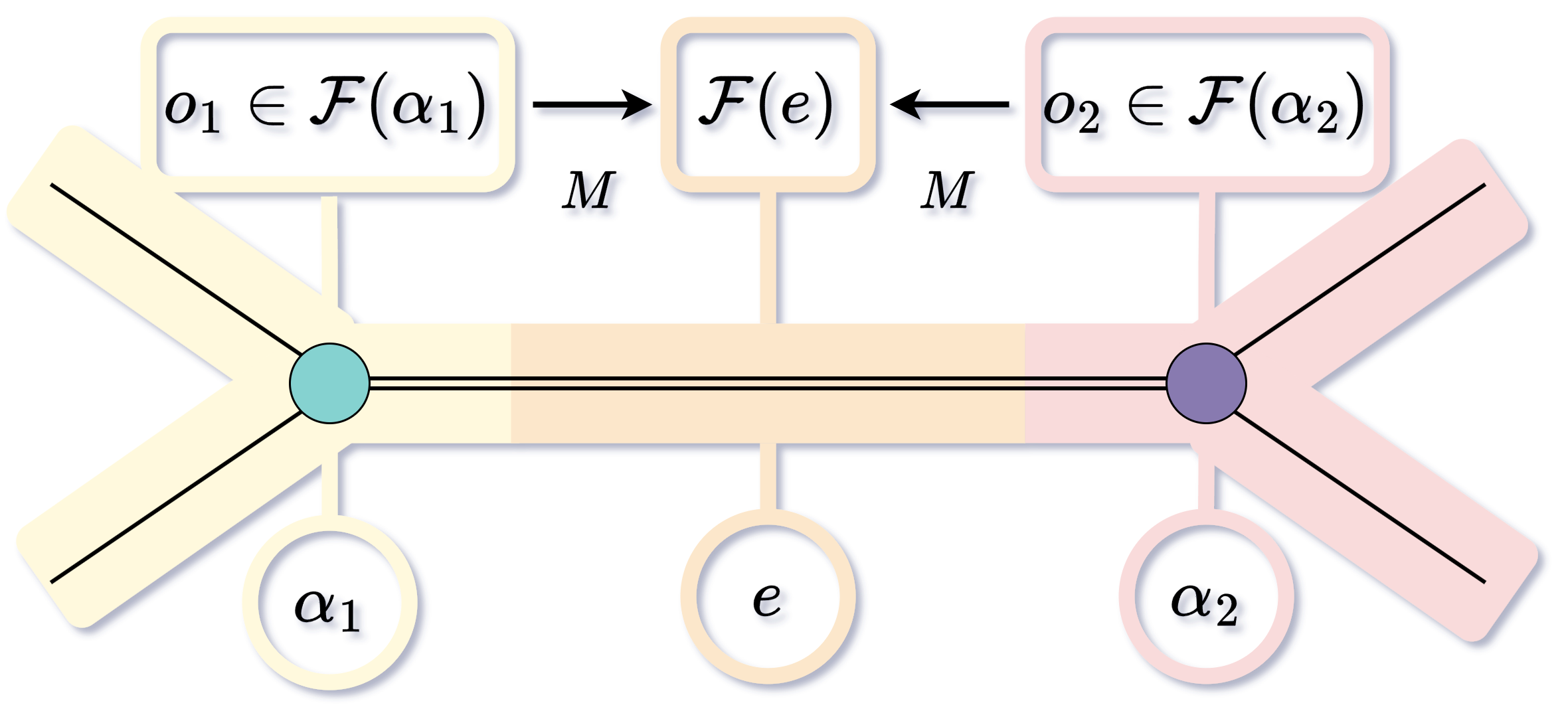}
  \caption{The observation vectors $o_1$ and $o_2$ of neighboring agents $\alpha_1$ and $\alpha_2$ are encoded into stalks $\mathcal{F}(\alpha_1)$ and $\mathcal{F}(\alpha_2)$, which are mapped onto the stalk of edge $e$ between them via restriction maps $M$.}
  \label{sheaf}
\end{figure}

\begin{table*}[!ht]
  \vspace{2mm}
  \centering
  \caption{Experimental Results. The symbol "$\uparrow$" indicates that a higher value is desirable, and vice versa.}
  \vspace{-2mm}
  \scalebox{0.75}{
    \input{tab_result}
  }
  \label{tab:results}
\end{table*}

As illustrated in Figure~\ref{sheaf}, \( o_1 \) and \( o_2 \) are the observation vectors of neighboring agents $\alpha_1$ and $\alpha_2$, and their corresponding stalks \( \mathcal{F}(\alpha_1) \) and \( \mathcal{F}(\alpha_2) \) should map to the same stalk $\mathcal{F}(e)$ on the edge $e$ via the restriction maps $M$, and they should match on the edge $\mathcal{F}(e)$ by definition of global sections, i.e., \( M (o_1) = M (o_2) \).
To satisfy these conditions and measure how close the current observations are to being within the space of global sections, we designed a self-supervise global section loss \( l_{\text{sec}} \). Specifically, \( l_{\text{sec}}^{\alpha_1} \) can be expressed as follows: 

\begin{equation}
    l_{{\rm{sec}}}^{{\alpha_1}} = \sum\limits_{\alpha_i \: {\rm in} \:\alpha_1\: \rm FOV } {{{\left| {M\left( {{o_i}} \right) - M\left( {{o_1}} \right)} \right|}^2}}
\end{equation}

\begin{equation}
    l_{\text{sec}} = \frac{1}{n} \sum_{i=1}^{n} l_{\text{sec}}^{\alpha_i}
\end{equation}

Where \( o_i \) represents the observation vectors of agent $\alpha_i$ within the FOV of agent $\alpha_1$, and \( M \) denotes the restriction map, $n$ is the number of agents. The function \( l_{\text{sec}} \) sums the mapping discrepancies between stalks, and by minimizing \( l_{\text{sec}} \), we ensure that the global section conditions are satisfied. 
When training converges, the global section loss is minimized, fulfilling the conditions set by sheaf theory, and ensuring that the agents achieve \textbf{consensus}.

We incorporate \(l_{\text{sec}}\) into the network 
updating to assist the agent in making decisions that adhere to the section conditions. As a result, the learned consensus among agents should be considered during learning the advantage value, we include \(M(s)\) in the advantage function to enable better evaluation of actions by the agents. The resulting Q-function in SIGMA is expressed as follows:

\begin{equation}
    Q(s, a) = V_{\theta_1}(s) +  A_{\theta_2}(s', a) - \frac{1}{|\mathcal{A}|} \sum_{a' \in \mathcal{A}} A_{\theta_2}(s', a')
\end{equation}
\begin{equation}
    s'=[M(s), s]
\end{equation}

\begin{equation}
    \mathcal{L}_{SIGMA} = \mathcal{L_Q} + \Lambda  \cdot {l_{{\rm{sec}}}}
\end{equation}


Here, \( Q(s, a) \) represents the Q value for action \(a\) in state \(s\), \( V_{\theta_1}(s) \) denotes the value function of state \(s\), which is parameterized by \(\theta_1\), and \( A_{\theta_2}(s', a) \) signifies the advantage function of action \(a\) in the enhanced state \(s'\), parameterized by \(\theta_2\). 
The term \(|\mathcal{A}|\) represents the size of the action space, \(\mathcal{L}_{SIGMA}\) is the loss function, and \(\Lambda\) is a hyperparameter.

\section{EXPERIMENTS}
In this section, we evaluate SIGMA through comprehensive simulation experiments, comparing its performance with SOTA baselines. We also conduct ablation studies to assess the impact of each component of our approach. Additionally, we test the robustness of the trained model by deploying it in simulation and real world environment.

\subsection{Main Results Comparison}

For our experiments, we train our models on structured environments of varying sizes, randomly chosen from a uniform distribution between 10 and 40, while consistently deploying 5 agents. During testing, we explore environments of 20, 40, and 60 sizes, scaling the number of agents from 4 up to 128.

Our evaluation include a comparison with 6 SOTA MAPF solutions: PRIMAL~\cite{primal}, MAPPER~\cite{mapper},
DHC~\cite{dhc}, DCC~\cite{dcc}, SCRIMP~\cite{scrimp}, and ALPHA~\cite{alpha}. Additionally, we benchmark against the searchbased, bounded-optimal centralized planner ODrM* with an inflation factor of \(\epsilon=2.0\)~\cite{odrm}. For a fair comparison, each planner was tested on the same set of 200 randomly-generated environments.

We employ three metrics to assess performance: 1) \textbf{Episode Length(EL)}: This measures the efficiency of a solution by counting the number of actions agents take to reach their goals within a single episode. 2) \textbf{Arrival Rate(AR)}: This is the percentage of agents that reach their goals across all episodes. 3) \textbf{Success Rate(SR)}: This metric evaluates a planner’s ability to completely fulfill a task. Notably, learning-based methods may show a low SR but still have a high AR, which highlights the importance of AR in evaluating episodes that nearly reach completion without being deemed total failures. The results are presented in Table~\ref{tab:results}.

In our experiments, SIGMA consistently outperforms other learning-based planners in terms of SR across all tasks. Notably, as the number of agents increases, SIGMA's improvement in SR significantly exceeds that of the baseline planners. For instance, in complex scenarios where most learning-based planners struggle or fail to solve the problems, such as with 128 agents on a 40$\times$40 map, SIGMA achieves a SR of 69\%. Even more impressively, on a larger 60$\times$60 map, SIGMA's SR reaches 80\%. These results highlight the effectiveness of our approach where the agents, through achieving consensus, successfully avoid congestion and overcrowding, demonstrating robust performance even under challenging conditions. Additionally, it is noteworthy that on a smaller 20x20 map, although SIGMA exhibits a high SR, the AR isn't as impressive. This indicates that while consensus effectively prevents congestion, it does not necessarily aid in navigating out of such congested scenarios efficiently.

\subsection{Ablation Analysis}

Our method focuses on encoding the sheaf structure(stalks and restriction maps) and integrating global section loss. These elements are applied to both the input of the advantage function and the loss function for updating the network to ensure the correctness of the sheaf structure. To analyze the importance of these elements, we experimented with three ablation variants of SIGMA: 1) \textbf{Encoded Stalk(ES)}: This variant tested only the stalks' encoding effectiveness by removing global section loss and restriction maps. 2) \textbf{Weighted Penalty(WP)}: This setup assessed the influence of global section loss within the loss calculation by excluding them. 3) \textbf{Feature Impact(FI)}: We evaluated the impact of removing restriction maps from the advantage function while keeping other elements constant.

\begin{figure}[thpb]
  \centering
  \includegraphics[scale=0.35]{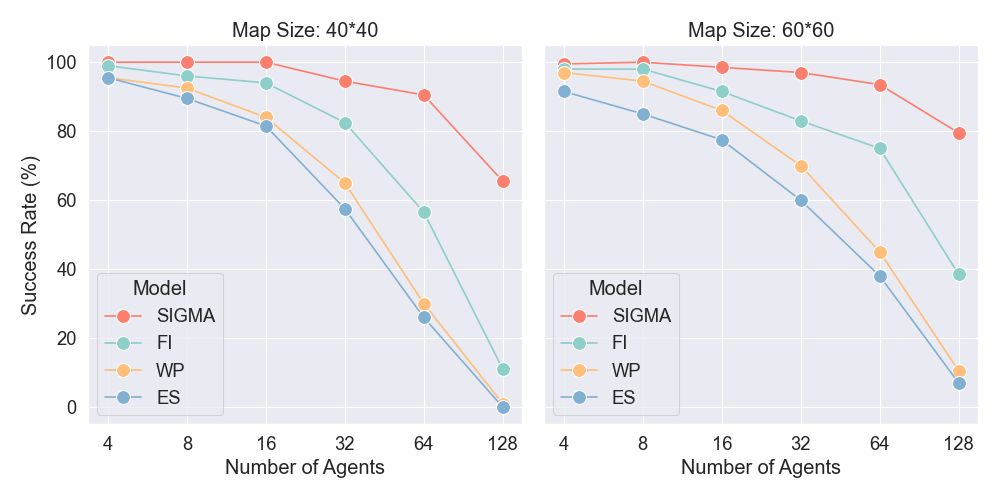}
  \caption{Success rates of ablation variants on 40x40 and 60x60 maps.}
  \label{ablation}
\end{figure}

The results of the ablation variants are shown in Figure~\ref{ablation}. Encoded stalk results in performance similar to the baselines, indicating that this component alone doesn't significantly influence success rates. The introduction of the restriction maps leads to a slight improvement in success rates, suggesting a beneficial but limited role in the overall system's performance. A notable enhancement is observed with the incorporation of global section loss, particularly as the number of agents increases. This enhancement significantly boosts success rates, demonstrating that global section loss effectively guide agents towards explicit consensus. When all components of the SIGMA framework are utilized, including both stalks, restriction maps, and global section loss, there is a further increase in performance. This indicates that the full integration of the sheaf structure is crucial for achieving consensus.

\subsection{Experimental Validation}
\begin{figure}[ht]
     \centering
     \begin{subfigure}[b]{0.22\textwidth}        
         \includegraphics[width=\textwidth]{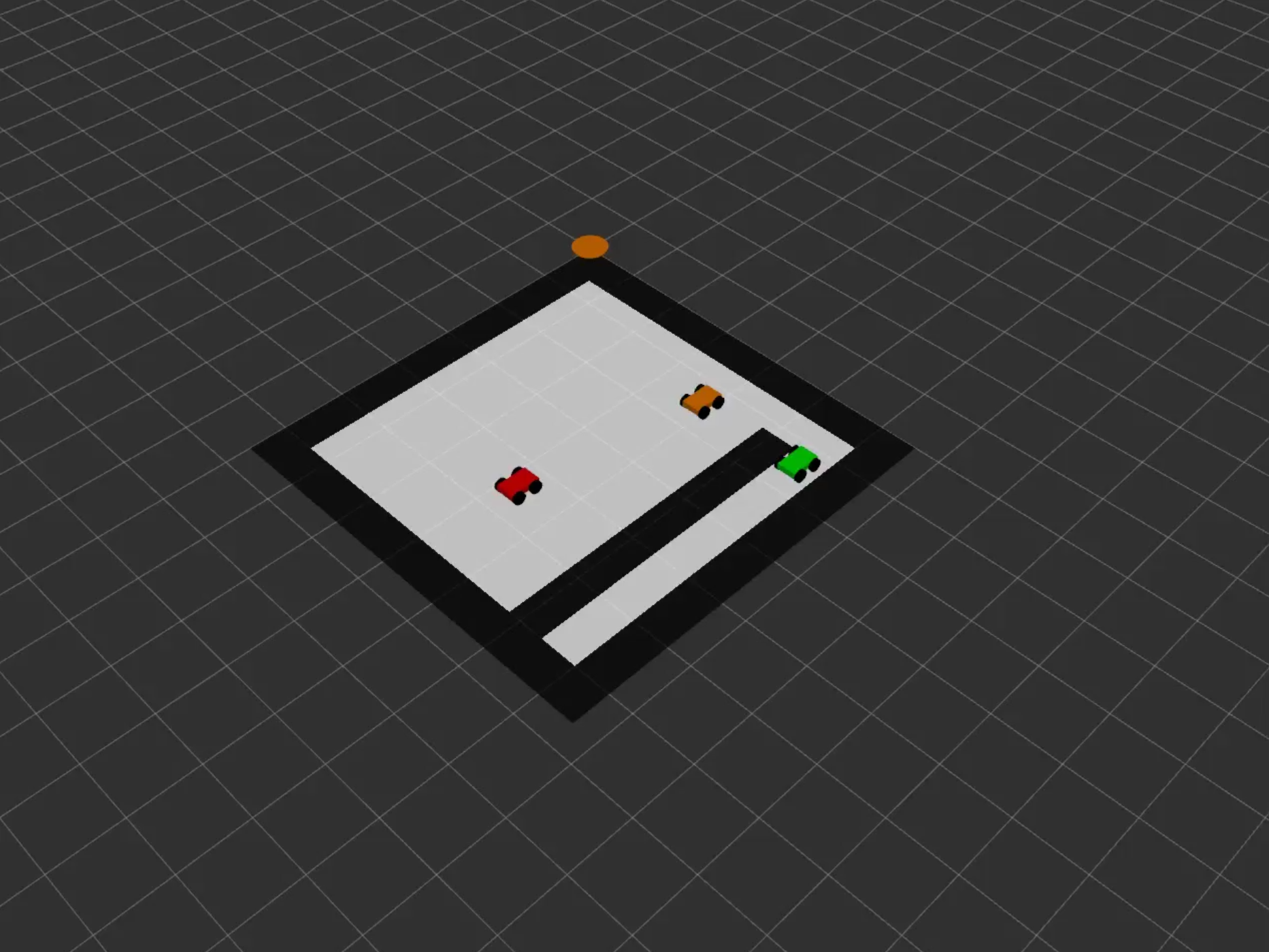}
         \caption{Simulation Environment}
         \label{sim}
     \end{subfigure}
     \hspace{5mm}
     \begin{subfigure}[b]{0.22\textwidth}       
         \includegraphics[width=\textwidth]{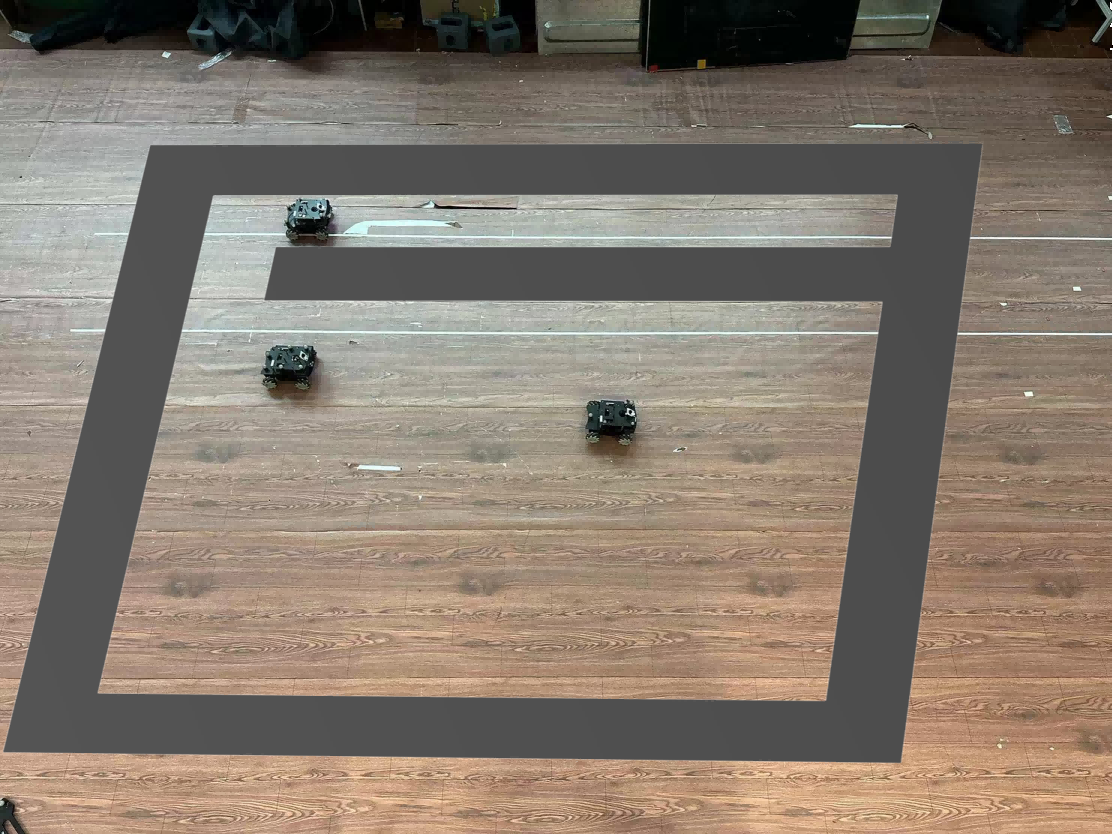}
         \caption{Real World}
         \label{real}
     \end{subfigure}
     \caption{Experiments with real robots on room-like map.}
     \label{expv}
\end{figure}

As shown in Figure \ref{expv}, Figure \ref{sim} represents the simulation environment, and Figure \ref{real} represents the real world environment. In this setup, each cell in the real world environment measures 0.3m on each side, slightly larger than the agents to ensure each agent occupies only one cell on the map. We employ 3 robots, each equipped with Mecanum wheels and measuring approximately 0.23m $\times$ 0.2m. The accurate positions of these robots are tracked using the \textit{OptiTrack Motion Capture System}. The starting and goal positions of the agents are randomly configured. In this experiment, although the robots recognize the virtual positions of obstacles and are programmed to avoid these areas, there are no physical obstacles in the real environment, preventing any interference with the line of sight needed for the OptiTrack motion capture system.

\section{CONCLUSIONS}

In this paper, we introduce SIGMA, a novel MAPF planner that pushes beyond the constraints of limited FOV commonly found in existing learning-based MAPF planners by utilizing sheaf theory to let all agents reason about and reach a team-wide consensus. Our approach efficiently encodes the sheaf structure within MAPF, incorporating global section loss to measure consistency. Through extensive experiments conducted on highly structured maps with varying team sizes and environmental complexities, SIGMA consistently outperforms SOTA learning-based MAPF planners and a bounded-optimal search-based planner in most scenarios. By rigorously proving the transition from local observation to global consensus, this work proposes a novel perspective to the MAPF research community.

In future work, we plan to focus on enriching the intricate relationships among agents by building consensus on more complex graphs. This will involve exploring advanced models and techniques that can handle and interpret the interactions and dependencies within larger, more complex network structures.







\section*{ACKNOWLEDGMENT}

This work was supported by the National Science and Technology Major Project(Grant No.2022ZD0116401), the National Natural Science Foundation of China(Grant No.62441617), and an Amazon Research Award.




\end{document}

%% file: tab_result.tex
\begin{tabular}{c|cccccc||cccccc||cccccc}
    \toprule[0.5mm]
    \multirow{2}{*}{\textbf{Method}} & \multicolumn{6}{c}{\textbf{EL}$\downarrow$} & \multicolumn{6}{c}{\textbf{AR}$\uparrow$} & \multicolumn{6}{c}{\textbf{SR}$\uparrow$}\\
    \cmidrule{2-19}

    & \multicolumn{18}{c}{\textbf{20 $\times$ 20 room-liked environment with 4, 8, 16, 32, 64, 128 agents}} \\
    
    \midrule\midrule    
    ODrM*  & 30.58  & 43.19  & 97.25  & 292.93 & 512.00 & 512.00 & 100\%   & 98.00\% & 88.00\% & 47.00\% &  0.00\% &  0.00\% & 100\% & 98\% & 88\% & 47\% &  0\% & 0\%  \\
    PRIMAL & 201.32 & 275.93 & 439.95 & 506.83 & 512.00 & 512.00 & 93.50\% & 90.88\% & 88.63\% & 81.72\% & 66.79\% & 35.74\% &  79\% & 67\% & 30\% &  2\% &  0\% & 0\%  \\
    MAPPER & 79.81  & 101.05 & 246.69 & 427.33 & 512.00 & 512.00 & 98.75\% & 97.53\% & 95.75\% & 89.71\% & 53.92\% &  6.96\% &  97\% & 97\% & 82\% & 41\% &  0\% & 0\%  \\
    DHC    & 45.50  & 73.86  & 175.22 & 354.43 & 509.69 & 512.00 & 99.00\% & 98.62\% & 96.56\% & 90.69\% & 69.70\% & 20.09\% &  98\% & 93\% & 77\% & 45\% &  1\% & 0\%  \\ 
    DCC    & 45.73  & 47.65 & 129.96 & 262.21 & 506.89 & - & 99.00\% & 99.50\% & 98.56\% & 95.91\% & 73.55\% & - & 97\% & 98\% & 86\% & 71\% & 4\% & - \\
    SCRIMP & 43.42  & 61.56  & 186.34 & 214.32 & 488.98 & --     & 99.25\% & 99.37\% & 98.87\% & 97.53\% & \textbf{82.32\%} & --      &  98\% & 96\% & 93\% & 75\% & 15\% & --   \\
    ALPHA  & \textbf{37.39}  & 52.26  & 120.65 & 310.21 & 503.87 & 512.00 & \textbf{100\%}   & \textbf{100\%}   & \textbf{99.75\%} & \textbf{97.69\%} & 70.12\% & \textbf{25.71\%} & \textbf{100\%} & \textbf{100\%} & 96\% & 78\% &  8\% & 0\%  \\
    \midrule [0.5mm]                                                                        
    \textbf{SIGMA}  & 38.04  & \textbf{40.54}  & \textbf{66.59} & \textbf{136.56} & \textbf{398.26} & 512.00 & \textbf{100\%} & \textbf{100\%} & 99.12\% & 96.75\% & 61.62\% & 8.46\% & \textbf{100\%} & \textbf{100\%} & \textbf{98\%} & \textbf{92\%} & \textbf{39\%} & 0\%  \\
    \midrule[0.5mm]
    & \multicolumn{18}{c}{\textbf{40 $\times$ 40 room-like environment with 4, 8, 16, 32, 64, 128 agents}} \\
    \midrule\midrule
    ODrM*  & 56.73  & 69.34  & 91.89  & 146.88 & 375.37 & 512.00 & 100\%   & 100\%   & 97.00\% & 85.00\% & 32.00\% &  0.00\% & 100\% &100\% & 97\% & 85\% & 32\% & 0\% \\
    PRIMAL & 285.55 & 384.93 & 463.86 & 492.82 & 511.80 & 512.00 & 91.00\% & 87.62\% & 85.56\% & 82.69\% & 73.14\% & 61.71\% &  73\% & 47\% & 23\% & 11\% &  1\% & 0\% \\
    MAPPER & 104.82 & 157.12 & 218.35 & 348.95 & 491.58 & 512.00 & \textbf{100\%}   & 99.37\% & 98.00\% & 93.71\% & 76.60\% & 51.02\% & \textbf{100\%} & 96\% & 91\% & 66\% & 16\% & 0\% \\
    DHC    & 104.19 & 127.78 & 188.62 & 263.81 & 427.02 & 512.00 & 97.75\% & 98.00\% & 97.88\% & 95.94\% & 91.17\% & 72.53\% &  92\% & 91\% & 80\% & 65\% & 28\% & 0\% \\
    DCC    & 63.25 & 102.40 & 142.88 & 201.43 & 338.16 & - & 99.75\% & 99.25\% & 99.12\% & 98.72\% & 96.47\% & - & 99\% & 95\% & 89\% & 85\% & 58\% & - \\
    SCRIMP & 58.53  & 91.84  & 116.05 & 183.54 & 396.93 & 484.76 & \textbf{100\%}   & 99.62\% & 99.56\% & 99.21\% & 94.10\% & 85.09\% & \textbf{100\%} & 97\% & 95\% & 84\% & 42\% & 12\% \\
    ALPHA  & 64.04  & 88.75  & 140.96 & 206.85 & 392.23 & 506.48 & \textbf{100\%}   & \textbf{100\%}   & 99.75\% & 99.34\% & 93.46\% & 73.99\% & \textbf{100\%} &\textbf{100\%} & 97\% & 93\% & 60\% & 7\% \\
    \midrule[0.5mm]
    \textbf{SIGMA}  & \textbf{57.23} & \textbf{77.10} & \textbf{88.33} & \textbf{128.70} & \textbf{223.73} & \textbf{380.48} & \textbf{100\%} & \textbf{100\%} & \textbf{100\%} & \textbf{99.56\%} & \textbf{98.50\%} & \textbf{92.11\%} & \textbf{100\%} & \textbf{100\%} & \textbf{100\%} & \textbf{95\%} & \textbf{89\%} & \textbf{69\%} \\
    \midrule[0.5mm]
    & \multicolumn{18}{c}{\textbf{60 $\times$ 60 room-liked environment with 4, 8, 16, 32, 64, 128 agents}} \\
    \midrule\midrule
    ODrM*  & 84.71  & 98.43  & 106.46 & 163.53 & 228.95 & 457.17 & 100\%   & 100\%   & 99.00\% & 88.00\% & 72.00\% & 14.00\% & 100\% &100\% & 99\% & 88\% & 72\% & 14\% \\
    PRIMAL & 363.45 & 465.35 & 495.85 & 508.17 & 512.00 & 512.00 & 84.75\% & 78.37\% & 79.75\% & 73.62\% & 71.51\% & 62.83\% &  54\% & 25\% & 11\% &  3\% &  0\% & 0\%  \\
    MAPPER & 177.61 & 241.31 & 280.69 & 388.55 & 490.02 & 512.00 & 99.50\% & 97.75\% & 98.31\% & 93.87\% & 85.96\% & 62.47\% &  97\% & 89\% & 90\% & 61\% & 17\% & 0\%  \\
    DHC    & 131.59 & 203.71 & 186.66 & 323.19 & 406.40 & 496.70 & 97.75\% & 96.75\% & 98.88\% & 95.16\% & 93.30\% & 87.79\% &  91\% & 77\% & 86\% & 54\% & 35\% & 7\%  \\
    DCC    & 113.64 & 145.24 & 170.38 & 268.39 & 331.01 & - & 98.25\% & 99.00\% & 98.88\% & 97.75\% & 95.50\% & - & 95\% & 93\% & 89\% & 69\% & 58\% & - \\
    SCRIMP & 106.79 & 166.37 & 125.50 & 211.03 & 421.65 & 498.72 & 99.50\% & 99.25\% & 99.61\% & 98.73\% & 96.79\% & 88.08\% &  98\% & 95\% & 97\% & 81\% & 31\% & 8\%  \\
    ALPHA  & 110.82 & 158.59 & 173.02 & 263.74 & 357.17 & 485.23 & 99.50\% & 99.25\% & 99.75\% & 99.03\% & 97.91\% & 89.16\% &  98\% & 97\% & 97\% & 86\% & 67\% & 25\% \\
    \midrule[0.5mm]
    \textbf{SIGMA}  & \textbf{91.02} & \textbf{103.63} & \textbf{109.53} & \textbf{147.76} & \textbf{187.96} & \textbf{332.35} & \textbf{100\%} & \textbf{100\%} & \textbf{100\%} & \textbf{100\%} & \textbf{99.78\%} & \textbf{98.52\%} & \textbf{100\%} & \textbf{100\%} & \textbf{100\%} & \textbf{98\%} & \textbf{96\%} & \textbf{80\%} \\
    \bottomrule[0.5mm]
\end{tabular}